\documentclass[final]{cvpr}
\pdfoutput=1

\usepackage{times}
\usepackage{epsfig}
\usepackage{graphicx}
\usepackage{amsmath}
\usepackage{amssymb}
\usepackage{adjustbox}
\usepackage{diagbox}
\usepackage{multirow}
\usepackage{hhline}
\usepackage{float}

\usepackage[pagebackref=true,breaklinks=true,colorlinks,bookmarks=false]{hyperref}

\def\confYear{CVPR 2021}

\newcommand{\mypara}{\vspace*{-3mm}\paragraph}
\newcommand{\myitem}{\vspace*{-3mm}\item}
\newcommand\blfootnote[1]{%
  \begingroup
  \renewcommand\thefootnote{}\footnote{#1}%
  \addtocounter{footnote}{-1}%
  \endgroup
}

\begin{document}

\title{3DIoUMatch: Leveraging IoU Prediction\\
for Semi-Supervised 3D Object Detection}

\author{He Wang\textsuperscript{1*} \quad Yezhen Cong\textsuperscript{2*} \quad Or Litany\textsuperscript{3} \quad Yue Gao\textsuperscript{2}  \quad Leonidas J.~Guibas\textsuperscript{1}\\
\textsuperscript{1}Stanford University \quad  \textsuperscript{2}Tsinghua University \quad \textsuperscript{3}NVIDIA\\}

\maketitle

\begin{abstract}
3D object detection is an important yet demanding task that heavily relies on difficult to obtain 3D annotations. To reduce the required amount of supervision, we propose 3DIoUMatch, a novel semi-supervised method for 3D object detection applicable to both indoor and outdoor scenes.
We leverage a teacher-student mutual learning framework to propagate information from the labeled to the unlabeled train set in the form of pseudo-labels. 
However, due to the high task complexity, we observe that the pseudo-labels suffer from significant noise and are thus not directly usable. 
To that end, we introduce a confidence-based filtering mechanism, inspired by FixMatch. 
We set confidence thresholds based upon the predicted objectness and class probability to filter low-quality pseudo-labels.
While effective, we observe that these two measures do not sufficiently capture localization quality. 
We therefore propose to use the estimated 3D IoU as a localization metric and set category-aware self-adjusted thresholds to filter poorly localized proposals.
We adopt VoteNet as our backbone detector on indoor datasets while we use PV-RCNN on the autonomous driving dataset, KITTI.
Our method consistently improves state-of-the-art methods on both ScanNet and SUN-RGBD benchmarks by significant margins under all label ratios (including fully labeled setting). 
For example, when training using only 10\% labeled data on ScanNet, 3DIoUMatch achieves 7.7\% absolute improvement on mAP@0.25 and 8.5\% absolute improvement on mAP@0.5 upon the prior art. 
On KITTI, we are the first to demonstrate semi-supervised 3D object detection and our method surpasses a fully supervised baseline from 1.8\% to 7.6\% under different label ratios and categories.

\end{abstract}

\vspace{-3mm}
\section{Introduction}
\blfootnote{*: equal contribution}
\blfootnote{Project page:~\href{http://THU17cyz.github.io/3DIoUMatch}{http://THU17cyz.github.io/3DIoUMatch}}
Object detection is a key task in 3D scene understanding. It provides a concise representation of raw sensor measurements in the form of semantically meaningful 3D bounding boxes. This low-dimensional representation can already serve numerous applications in autonomous driving and AR/VR, as well as in robot navigation and manipulation.  As a result, in recent years there has been a surge of interest in developing improved object detection pipelines and indeed current state-of-the-art methods show impressive performance. Yet, much of their success is attributed to the availability of large datasets of 3D scenes that are carefully annotated. While rapid advances in sensor technology facilitate the collection of 3D scenes at scale, annotating them remains the main bottleneck. This calls for detection methods that can leverage both labeled and unlabeled data at train time. 

In this work, we aim to address this requirement by proposing a novel semi-supervised 3D object detection method which we dub 3DIoUMatch. 
As a generally applicable method, 3DIoUMatch can be applied to both indoor scene datasets, \textit{i.e.} ScanNet\cite{dai2017scannet} and SUN-RGBD\cite{song2015sun}, and outdoor datasets,  \textit{i.e.} KITTI\cite{geiger2013vision}. We adopt popular point-based object detectors, VoteNet~\cite{qi2019deep} and PV-RCNN~\cite{shi2020pv}, as our backbone object detection networks for the indoor and outdoor scenes, correspondingly.
To provide supervision to the unlabeled scenes, we leverage a teacher-student mutual learning framework~\cite{tarvainen2017mean} and use the bounding box predictions from the teacher network as pseudo-labels to supervise the student network on unlabeled data. 
However, unlike most pseudo-label techniques that were designed for classification, in the highly complex (joint regression and classification) task of object detection, we observe that the pseudo-labels suffer from significant noise, and using them directly is suboptimal.  

Inspired by FixMatch~\cite{sohn2020fixmatch}, the state-of-the-art semi-supervised learning (SSL) method for 2D image classification that proposed confidence-based filtering to improve pseudo-label quality, we adopt a pseudo-label filtering mechanism for 3D object detection by setting thresholds on predicted class probabilities (and objectness scores for VoteNet), so as to filter out teacher proposals with potentially erroneous semantic labels or ones not belong to foreground. 
While effective, these criteria alone are not sufficient to capture localization quality, and the pseudo-labels may still have large errors in the bounding box parameters. To that end, we further propose to leverage estimated IoU (intersection over union) as a localization quality measure for pseudo-label filtering. IoU estimation was first proposed in the context of 2D object detection as a localization confidence in the pioneering work IoU-Net~\cite{jiang2018acquisition}, where estimated IoU was proven successful in replacement of class confidence for test-time Non-Maximal Suppression (NMS).
To the best of our knowledge, leveraging IoU estimation for pseudo-label filtering is a novel idea for SSL on both 2D and 3D object detection. 
Equipping the detectors with a 3D IoU estimation module, we are able to filter out poorly localized pseudo-labels and leverage estimated IoU for both train-time and test-time NMS.

A key challenge when filtering based on IoU estimation is how to properly set the threshold. Unlike objectness and class confidence for which high threshold values (e.g. 0.9) work well, 3D IoU is more sensitive to small errors. Setting the threshold too high would reduce the number of pseudo-labels to very few, from which little could be learned. To balance between quality and coverage, we propose a two-stage filtering process: first, using a relatively low IoU threshold; then, an IoU-guided class-aware \textit{Lower-Half Suppression} (LHS) that removes only half of the highly-overlapping boxes with low predicted IoU. Our proposed LHS thus naturally sets a threshold that is both dynamic and class-aware. Our experiments show that LHS outperforms IoU-guided NMS, which suppresses all but the top one during semi-supervised training. 

Our method consistently improves upon the previous state-of-the-art method, SESS~\cite{zhao2020sess}, on both ScanNet and SUN-RGBD benchmarks by significant margins. When using only 10\% labeled data on ScanNet, 3DIoUMatch outperforms SESS by 7.7 absolute improvement on mAP@0.25 and by 8.5 absolute improvement on mAP@0.5. When using 5\% labeled data on SUN-RGBD, 3DIoUMatch outperforms SESS by 4.8 absolute improvement on mAP@0.25 and by 8.0 absolute improvement on mAP@0.5. 
On KITTI, we are the first to demonstrate semi-supervised 3D object detection work and surpass fully-supervised baseline by large margins under all label ratios.

Our main contributions can be summarized as follows:
\begin{enumerate}
    \myitem We propose a novel semi-supervised method for 3D object detection in point clouds based on pseudo-label propagation along with a carefully designed filtering mechanism. 
    \myitem For the first time, we leverage predicted 3D IoU as a localization confidence score for pseudo-label filtering, and further propose IoU-guided Lower-Half Suppression for robust pseudo-label deduplication. This idea is generally applicable and can be coupled to different 3D detectors on both indoor and outdoor scenes.
    %
    \myitem We achieve markedly improved performance over the previous state-of-the-art semi-supervised 3D object detection methods on the two major indoor object detection benchmarks, ScanNet and SUN-RGBD, under low label ratios and fully labeled setting. As the first semi-supervised 3D object detection work on KITTI, we also achieve significant improvements compared to fully supervised method.
\end{enumerate}



\section{Related Works}
\begin{figure*}[t]
\begin{center}
   \includegraphics[width=0.8\linewidth]{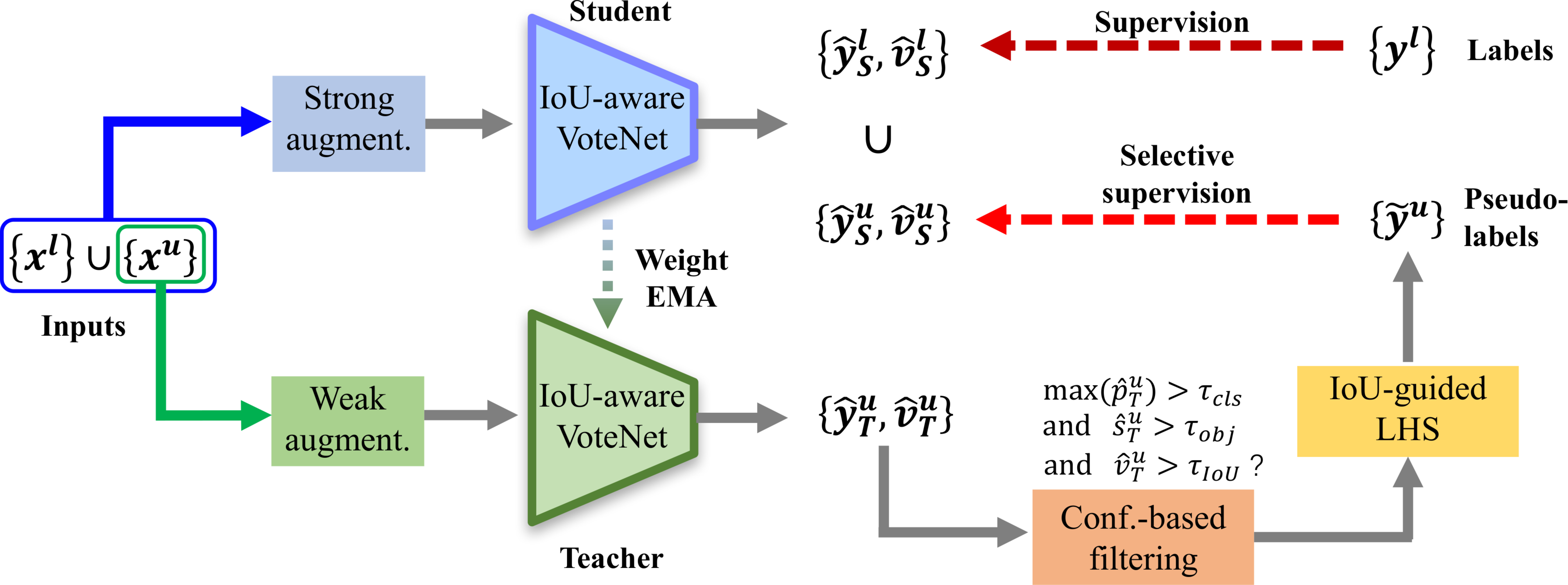}
\end{center}
   \caption{\textbf{3DIoUMatch pipeline at semi-supervised training stage.} We adopt as our backbone an extended version of VoteNet with an additional 3D IoU estimation module. For SSL, we utilize a teacher-student mutual learning framework, composed of a learnable student taking strongly augmented input data and an EMA teacher taking weakly augmented input samples. On labeled data, the student network is supervisedly trained. On unlabeled data, the student network takes pseudo-labels from its EMA teacher. To improve the quality of pseudo-label, we adopt a confidence-based filtering mechanism that filters out predictions that fail to pass all thresholds on class probability, objectness, and 3D IoU. We further use IoU-guided Lower-Half Suppression to remove the duplicated predictions. Using the filtered pseudo-labels, we selectively supervise the student predictions that are around the bounding boxes in the pseudo-labels.}
\label{fig:long}
\end{figure*}

\paragraph{Semi-Supervised Learning (SSL)}
Many of the recent SSL methods~\cite{berthelot2019mixmatch, xie2019unsupervised, berthelot2019remixmatch} leverage consistency regularization, first proposed in \cite{sajjadi2016regularization, laine2016temporal}, which enforces the model to predict consistently across label-preserving data augmentation of different intensity. Borrowing the concept from Mean Teacher~\cite{tarvainen2017mean}, the model with frozen weight can be viewed as the teacher model, otherwise student model. Some methods~\cite{berthelot2019mixmatch}, following Mean Teacher, make the teacher model as the EMA of the student model for further regularization. Pseudo labeling~\cite{lee2013pseudo} is another popular class of SSL method which can also be treated as a kind of consistency regularization, as one output of the unlabeled data is enforced to be consistent with the other (the pseudo-labels) by being supervised with the other. To improve the quality of pseudo-labels, FixMatch~\cite{sohn2020fixmatch}, a state-of-the-art SSL work on image classification, has shown that the student network can improve significantly by setting a classification confidence threshold $\tau_{cls}$ and filtering out low-confidence predictions from the teacher. With the filtered pseudo-labels, the student model only gets supervised on the unlabeled data whose pseudo-labels are kept. Another key factor to the success of these methods is strong data augmentation. It has been shown crucial to many SSL works~\cite{sajjadi2016regularization, laine2016temporal, xie2019unsupervised}. Recent works~\cite{berthelot2019remixmatch, sohn2020fixmatch} proposed to adopt even more powerful augmentation such as RandAugment~\cite{cubuk2020randaugment} and Cutout~\cite{devries2017improved}.

\mypara{Semi-Supervised Object Detection}
Since the beginning of the deep learning era, tremendous progress has been made in 2D object detection, e.g region-based detectors~\cite{girshick2014rich, girshick2015fast, ren2015faster} and single-stage detectors~\cite{liu2016ssd, redmon2016you, tian2019fcos}. Similarly in 3D object detection, a number of deep learning methods have been proposed for different 3D data modalities, e.g. RGBD-based detectors~\cite{qi2018frustum,qi2020imvotenet}, point-based detectors~\cite{yi2019gspn, shi2019pointrcnn, lang2019pointpillars, qi2019deep}, voxel-based detectors~\cite{zhou2018voxelnet}, point-voxel-based detectors~\cite{shi2020pv}, etc.

Despite the great progress in both 2D and 3D object detection, most works focused on a fully-supervised setting. A few works~\cite{hoffman2014lsda, gao2019note} have proposed to leverage unlabeled data or weakly-annotated data for 2D object detection. 
Under a standard SSL setting as we follow, CSD~\cite{jeong2019consistency} proposed a consistency regularization method to enforce the consistency between predictions from an image and its flipped version. STAC~\cite{sohn2020simple} adopts a two-stage scheme for training Faster R-CNN~\cite{ren2015faster}: in the first stage it pre-trains a detector with labeled data only and then predicts the pseudo labels for the unlabeled data; in the second stage, STAC leverages asymmetric data augmentation and the pseudo-label filtering mechanism to remove object proposals with low confidence. Note that the pseudo-labels are only generated once at the end of the first stage. 

The only prior work on semi-supervised point-based 3D object detection, is SESS~\cite{zhao2020sess}. SESS is built upon VoteNet~\cite{qi2019deep} and adopts a two-stage training scheme. It leverages a mutual learning framework composed of an EMA teacher and a student, uses asymmetric data augmentation, and enforces three kinds of consistency losses between the teacher and student outputs. Although SESS brings noticeable improvements upon a vanilla VoteNet when using only a small portion of labeled data, we find their consistency regularization suboptimal, as it is uniformly enforced on all the student and  teacher predictions. In this work, we instead propose to apply confidence-based filtering to improve the quality of pseudo-labels from the teacher predictions and we are the first (in both 2D and 3D object detection) to introduce IoU estimation for localization filtering. 

\mypara{IoU Estimation}
IoU estimation was first proposed in a 2D object detection work IoU-Net~\cite{jiang2018acquisition}, which proposed an IoU head that runs in parallel to bounding box refinement and is differentiable w.r.t. bounding box parameters. IoU-Net adds an IoU estimation head to several off-the-shelf 2D detectors and uses IoU estimation instead of classification confidence to guide NMS, which improves the performance consistently over different backbones. Thanks to its differentiability, IoU-Net can perform IoU optimization on bounding box parameters for iterative refinement, which further brings noticeable performance improvement.


For 3D object detection, STD~\cite{yang2019std} follows IoU-Net to add a simple IoU estimation branch parallel with the box estimation branch and to guide NMS with IoU estimation. 
PV-RCNN~\cite{shi2020pv} devises a similar 3D IoU estimation module and use it at IoU-guided NMS stage. These two modules, unfortunately, are not suitable for IoU optimization as the features fed to the IoU estimation branch are not differentiable w.r.t. the bounding box size.
Since the original VoteNet is not equipped with an IoU module, we devise a differentiable point-cloud-based 3D IoU estimation module is simple yet effective that can support the IoU optimization.

\section{Method}
In this section, we describe our solution in detail. We first formulate our problem in \ref{sec:formulation} and then summarize the two object detection backbones, PV-RCNN and VoteNet, in \ref{sec:votenet}. We use VoteNet as an example to illustrate our proposed 3DIoUMatch pipeline in \ref{sec:pipeline}. We further explain how we use the estimated 3D IoU for pseudo-label filtering and deduplication in \ref{sec:pseudo-label}. Finally, we illustrate how we leverage the pseudo-labels for supervision in \ref{sec:supervise}.

\subsection{Problem Definition}\label{sec:formulation}
Given a 3D point cloud representation of a scene $\boldsymbol{x} \in \mathbb{R}^{N\times 3}$ containing a set of objects $O = \{o^{(j)}\}$, we aim at detecting the amodal oriented 3D bounding boxes of all objects in $O$, along with their semantic class labels. In particular, we are interested in accomplishing this task under challenging conditions of limited supervision
%
%
where we have access to a (small) set of labeled scenes $\{\boldsymbol{x}^l_i, \boldsymbol{y}^l_i\}_{i=1}^{N_l}$ and a set of unlabeled scenes $\{\boldsymbol{x}^u_i\}_{i=1}^{N_u}$, where $N_l$ and $N_u$ are the number of labeled and unlabeled scenes, respectively. For a labeled scene $\boldsymbol{x}$, the label $\boldsymbol{y}$ comprises bounding box parameters $\{\boldsymbol{b}^{(j)}\}$ and semantic class labels $\{q^{(j)}\}$ of all ground truth objects $\{o^{(j)}\}$.

\subsection{IoU-aware 3D Object Detection}
\label{sec:votenet}
We experiment our SSL method on two 3D detectors, VoteNet \cite{qi2019deep} and PV-RCNN \cite{shi2020pv}. VoteNet is a single-stage indoor point cloud detector while PV-RCNN is a two-stage outdoor point cloud detector. They both take point clouds only for inputs and output a list of bounding boxes after Non-Maximum Suppression (NMS) for each scene, which contain the prediction of center, size, orientation and semantic class. However, their architecture are very different partly due to the great discrepancy between indoor and outdoor scenes.

\noindent\textbf{Indoor scene detector: VoteNet} VoteNet \cite{qi2019deep} is built upon PointNet++~\cite{qi2017pointnet++} backbone, and first processes the input point cloud $\{x_i\}_{i=1}^{N}$ to generate a sub-sampled set of $M < N$ seed points enriched with high-dimensional features $\{[x_i;f_i]\in\mathbb{R}^{3+C}\}_{i=1}^{M}$.
Next, each seed point votes for the center of the object it belongs to, and the votes are grouped into $K$ clusters. Finally, each of the K vote clusters is aggregated to make a prediction of a 3D bounding box parameters $\boldsymbol{b}^{(k)}$, a corresponding objectness score $s_k = \text{P}(\boldsymbol{b}^{(k)}~ \text{is an object})$, 
and a probability distribution $\{p_{cls}\}$ over $L$ possible semantic classes. 
The bounding box parameters $\boldsymbol{b}$ are its center location $\mathbf{c} \in \mathbb{R}^{3}$, scale $\mathbf{d}\in \mathbb{R}^{3}$, and orientation $\theta$ around the upright axis.

At train time, VoteNet jointly minimizes a weighted combination of the following target losses: vote coordinate regression, objectness score binary classification, box center regression, bin classification and residual regression for heading angle, scale regression, and category classification. At test time, VoteNet applies Non-Maximum Suppression (NMS) based on objectness score to remove duplicated bounding boxes. Here, we instead rely on a 3D IoU estimation module designed for VoteNet. For more details, refer to the supplementary materials.

\noindent\textbf{Outdoor scene detector: PV-RCNN} PV-RCNN\cite{shi2020pv} is a high-performance and efficient LiDAR point cloud detector that deeply integrates both 3D voxel CNNs and PointNet++-style set abstraction to learn more discriminative point cloud features. Specifically, PV-RCNN first passes the 3D scene through a novel voxel set abstraction module based on sparse 3D CNN to get a set of keypoints with representative scene features. Then RoI grid pooling is then applied to the keypoints to abstract proposal-specific features into RoI grid points. The RoI grid points containing rich context information are finally used to accurately estimate bounding box parameters.

PV-RCNN itself incorporates an IoU-estimation module which can predict the IoU of each bounding box and use it to guide the sorting of the boxes.

\subsection{3DIoUMatch for SSL on 3D object detection}\label{sec:pipeline}
We take VoteNet as our example and our method with PV-RCNN is similar.
With the incorporation of 3D IoU module into VoteNet, we construct an IoU-aware VoteNet for SSL on 3D object detection. Our proposed solution is comprised of two training stages:  a pre-training stage, where we train our IoU-aware VoteNet on the labeled data, followed by an SSL stage where the entire data is utilized by pseudo-labeling the unlabeled scenes.

\mypara{Pre-training.} We start by training our IoU-aware VoteNet in a supervised manner, using the labeled set $\{\boldsymbol{x}^l_i, \boldsymbol{y}^l_i\}_{i=1}^{N_l}$. The training loss is a sum over the original VoteNet losses $L_\text{votenet}$ and 3D IoU loss $L_\text{IoU}$. Once converged, we clone the network to create a pair of student and teacher networks. 

\mypara{Semi-supervised training through a teacher-student framework.}
We follow a teacher-student mutual learning framework~\cite{tarvainen2017mean} and train our networks on both labeled $\{\boldsymbol{x}^l_i, \boldsymbol{y}^l_i\}_{i=1}^{N_l}$ and unlabeled data $\{\boldsymbol{x}^u_i\}_{i=1}^{N_u}$. Each training batch contains a mixture of $\{\boldsymbol{x}^l_i\}_{i=1}^{B_l}$ labeled samples and $\{\boldsymbol{x}^u_i\}_{i=1}^{B_u}$ unlabeled samples. 

For labeled samples, we supervise the student network using ground truth supervisions (as done in the pre-training stage) whereas for unlabeled samples, the student networks is supervised using pseudo-labels $\{\boldsymbol{\tilde{y}}^u_i\}_{i=1}^{N_u}$ generated
from the teacher network. The final loss is formed as:
$$ L = L_{l}(\{\boldsymbol{x}^l_i\}_{i=1}^{N_l}, \{\boldsymbol{y}^l_i\}_{i=1}^{N_l}) + \lambda_{u}L_{u}(\{\boldsymbol{x}^u_i\}_{i=1}^{N_u},\{\boldsymbol{\tilde{y}}^u_i\}_{i=1}^{N_u}) $$ where $\lambda_{u}$ is the unsupervised loss weight.

To succeed in semi-supervised learning, it is crucial for the teacher network to generate high-quality pseudo-labels and maintain a reliable performance margin over the student network throughout the training. As commonly used in SSL literature, e.g. Mean Teacher~\cite{tarvainen2017mean} and SESS~\cite{zhao2020sess}, we adopt an EMA teacher.
We further leverage asymmetric data augmentation and pseudo-label filtering (see Sec.\ref{sec:pseudo-label}).

To be in a position of advantage, the teacher network takes input data with weak augmentation only while the student network uses stronger data augmentation. 
We share the same data augmentation strategy with SESS. The input point clouds to our teacher network are augmented only by random sub-sampling while the inputs to the student network further undergo a set of stochastic transformation $\mathcal{T}$, including random flip, random rotation around the upright axis, and a random uniform scaling.

\subsection{Pseudo-Label Filtering and Deduplication}\label{sec:pseudo-label}
In the teacher-student framework, the performance gap between the teacher and the student is usually quite marginal given that these two models are just different by EMA on weight and data augmentation strength. Hence, it is not always true that the teacher prediction is more accurate than the student's on a specific training sample. On unlabeled data, the student model will only benefit from the pseudo-labels that are more accurate than its predictions. Therefore we should filter out low-quality predictions from the teacher model and only supervise the student model with the rest of the teacher model predictions.


\mypara{Jointly filtering based on class, objectness, localization confidences}
For VoteNet, we propose to set an objectness threshold $\tau_{obj}$ and filter out bounding box predictions with objectness score $s < \tau_{obj}$. We further propose to set a classification confidence threshold $\tau_{cls}$ for filtering out predictions that are likely to contain a wrong class label. 

Note that none of these two confidence measures capture the accuracy of bounding box parameter predictions. We propose to predict a 3D IoU for each predicted bounding box, use the 3D IoU estimation as a localization confidence, and set a localization threshold $\tau_\text{IoU}$ to filter out poorly localized predictions. Formally, we remove all the predictions that fail to satisfy all three confidence thresholds, i.e. $s > \tau_{obj}$, $\max(p_{cls}) > \tau_{cls}$, and $v > \tau_\text{IoU}$. 

\mypara{IoU-guided lower-half suppression for deduplication}
After the confidence-based filtering, there is still a lot of duplicated bounding box predictions that may introduce harmful noise to our pseudo-labels. NMS is a standard process in object detection for duplicate removal before evaluation, which takes a set of highly overlapped bounding box predictions that share the same class prediction, ranks them according to a confidence score and removes all but the top-1 prediction. STAC~\cite{sohn2020simple} applies class confidence based NMS to teacher predictions during pseudo-label generation. 

The default NMS used in VoteNet is based on objectness confidence. Given that objectness score doesn't capture the localization quality, a train-time IoU-guided NMS will naturally perform better (see Table.\ref{tab:ablation}), where we use the product of predicted IoU and predicted objectness as the ranking metric.
However, using the top one selected by IoU-guided NMS can still be suboptimal, since the predicted IoU will inevitably carry some errors. We argue that different from the test time scenario, pseudo-labels do not need to be fully deduplicated. Imagine this situation: if a  bounding box predicted by the student is $0.2 m$ to the left of its corresponding ground truth, it is a foreground object and will get bounding box supervision in VoteNet. However, if unfortunately the pseudo-label survives after non-maximal suppression is to the right of the ground truth more than $0.1m$, this predicted bounding box may lose supervision and be treated as a background box. This example shows that strict non-maximal suppression can lead to a smaller number of student model predictions that can receive supervision. Since we cannot know the best pseudo label among a bunch of highly-overlapped ones, it's fine to be less strict. To this end, we propose a novel Lower-Half Suppression, or in short, LHS, that only discards half of the proposals with lower predicted IoU.  We argue that since LHS suppresses bounding boxes sharing the same class label, this suppression can be seen as a second-step class-aware self-adjusted filtering, which sets dynamic thresholds among the overlapping bounding boxes to keep the ones with higher confidence and hence find a better balance between pseudo-label quality and the amount of supervision. We also use the product of predicted IoU and predicted objectness as the confidence metric.


\mypara{Final-step pseudo-label processing}
After the filtering and IoU-guided LHS, we now have high-quality predictions $\{\hat{{y}}_T^u\}_{k=1}^{K'}$ from the teacher network, where $K'$ is the number of bounding boxes remains. Given that the student model inputs go through a stronger augmentation including an additional geometric transformation $\mathcal{T}$, in synchronize with the student model inputs, the bounding box parameters of the pseudo-labels need to go through the same transformation $\mathcal{T}$, namely $\tilde{{\boldsymbol b}}^u = \mathcal{T}(\hat{{\boldsymbol b}}_T^u)$. We further take convert the predicted class probability distribution $\hat{p}_T^u$ into semantic class label via $\tilde{q}^u = \max(\hat{p}_T^u)$. Now we obtain the filtered pseudo-labels $\tilde{y} = \{\tilde{\boldsymbol{b}}^u, \tilde{q}^u\}_{k=1}^{K'}$.

\newcommand{\tabincell}[2]{\begin{tabular}{@{}#1@{}}#2\end{tabular}}  

\begin{table*}[h]
\begin{tabular}{|c|c|c|c|c|c|c|c|c|c|}
\hline
& & \multicolumn{2}{c|}{5\%} & \multicolumn{2}{c|}{10\%} & \multicolumn{2}{c|}{20\%} & \multicolumn{2}{c|}{100\%} \\ 
  \cline{3-10} \multirow{-2}{*}{Dataset} & \multirow{-2}{*}{Model} & \tabincell{c}{mAP\\@0.25} & \tabincell{c}{mAP\\@0.5} & \tabincell{c}{mAP\\@0.25} & \tabincell{c}{mAP\\@0.5} & \tabincell{c}{mAP\\@0.25} & \tabincell{c}{mAP\\@0.5} & \tabincell{c}{mAP\\@0.25} & \tabincell{c}{mAP\\@0.5} \\ \hline

\multirow{5}{*}{ScanNet} & VoteNet  & 27.9±0.5 & 10.8±0.6 & 36.9±1.6 & 18.2±1.0 & 46.9±1.9 & 27.5±1.2 & 57.8 & 36.0 \\ \cline{2-10} 
& SESS reported     & $\backslash$ & $\backslash$ & 39.7±0.9 & 18.6 & 47.9±0.4 & 26.9 & 62.1 & 38.8  \\ \cline{2-10} 
                         & SESS     & 32.0±0.7 & 14.4±0.7 & 39.5±1.8 & 19.8±1.3 & 49.6±1.1 & 29.0±1.0 & 61.3 & 39.0  \\ \cline{2-10} 
                         & Ours & \textbf{40.0±0.9} & \textbf{22.5±0.5} & \textbf{47.2±0.4} & \textbf{28.3±1.5} & \textbf{52.8±1.2} & \textbf{35.2±1.1} & \textbf{62.9} & \textbf{42.1} \\ \cline{2-10} 
                         & Abs. improve. & +8.0 & +8.1 & +7.7 & +8.5 & +3.2 & +6.2 & +1.6 & +3.1 \\ \hline
\multirow{5}{*}{SUN-RGBD} & VoteNet   & 29.9±1.5 & 10.5±0.5 & 38.9±0.8 & 17.2±1.3 & 45.7±0.6 & 22.5±0.8 & 58.0 & 33.4      \\ \cline{2-10} 
& SESS reported     & $\backslash$ & $\backslash$ & 42.9±1.0 & 14.4 & 47.9±0.5 & 20.6 & 61.1 & 37.3  \\ \cline{2-10}
                         & SESS      & 34.2±2.0 & 13.1±1.0 & 42.1±1.1 & 20.9±0.3 & 47.1±0.7 & 24.5±1.2 & 60.5 & 38.1  \\ \cline{2-10} 
                         & Ours & \textbf{39.0±1.9} & \textbf{21.1±1.7} & \textbf{45.5±1.5} & \textbf{28.8±0.7} & \textbf{49.7±0.4} & \textbf{30.9±0.2} & \textbf{61.5} & \textbf{41.3} \\ \cline{2-10} 
                         & Abs. improve. & +4.8 & +8.0 & +3.4 & +7.9 & +2.6 & +6.4 & +1.0 & +3.2  \\ \hline

\end{tabular}
\vspace{1mm}

\caption{Comparison with VoteNet and SESS on ScanNet val set and SUN RGB-D val set under different ratios of labeled data. We report the mAP@0.25 and mAP@0.5 as mean±standard deviation across 3 runs under different random data splits. Due to the randomness of the data splits and our better pre-training protocol, SESS results provided by us are higher than those reported in the paper on mAP@0.5, and the mAP@0.25 results differ a little (the only difference is the pre-trained weights and data splits). The final improvement is the absolute improvement of our method over SESS results provided by us. Following SESS, we also report the results with 100\% labeled data, where we simply make a copy of the full dataset as unlabeled data and train our method.}
\label{tab:main}
\end{table*}

\subsection{Selective Supervision using Pseudo-Labels}\label{sec:supervise}

For our generated pseudo-labels, there is no guarantee that the labels can cover all the ground truth objects from $O$ due to the filtering and potentially inaccurate teacher predictions. Given the incompleteness of our filtered pseudo-labels, we are relatively confident about the bounding boxes in this set but student predictions far away from all of our pseudo-labels are not necessarily negative. Our experiments show that supervising objectness on unlabeled data using the pseudo-labels seriously hurts the performance. For similar reasons, we do not supervise vote loss, which is a unique element in VoteNet and not shown in other detectors. For more analysis and experimental proof for this, we refer the readers to the supplementary materials. 
In this case, we will only supervise the bounding boxes in the vicinity of the pseudo bounding boxes and aim to improve their bounding box quality. More specifically, we stick to the way how VoteNet select foreground objects for bounding box parameter supervision: we supervise bounding box parameters and class for a prediction only if the vote that generates this prediction is within $0.3 \text{m}$ of any bounding box in the pseudo-labels. For this set of pseudo-foreground predictions, we adopt the same way that VoteNet establishes association and enforce original VoteNet losses except for objectness loss and vote loss.

\section{Experiments}

\subsection{Datasets and Evaluation Metrics}

\paragraph{Indoor Datasets: ScanNet and SUNRGB-D}
We evaluate our VoteNet-based 3DIoUMatch on two major indoor datasets, ScanNet~\cite{dai2017scannet} and SUN RGB-D~\cite{song2015sun}. ScanNet is an indoor scene dataset consisting of 1513 reconstructed meshes, among which 1201 are training samples and the rest are validation samples. SUN RGB-D contains 10335 RGB-D images of indoor scenes which are split into 5285 training samples and 5050 validation samples. For both datasets, we follow~\cite{qi2019deep,zhao2020sess} for pre-processing data and labels to train our method and we report mAP@0.25 (mean average precision with 3D IoU threshold 0.25) and mAP@0.5 in the following experiments.

\paragraph{Outdoor Dataset: KITTI}
As for our PV-RCNN-based 3DIoUMatch, we use KITTI for evaluation. KITTI \cite{geiger2013vision} is a very popular dataset for autonomous driving which consists of fine annotations for 3D detection. There are 7481 outdoor scenes for training and 7518 for testing, and the training samples are generally divided into a train split of 3712 samples and a validation split of 3769 samples. We follow~\cite{shi2020pv} for data pre-processing and report the mAP with 40 recall positions, with a rotated IoU threshold 0.7, 0.5, 0.5 for the three classes, car, pedestrian, and cyclist, respectively.

\begin{table*}[t]
\centering
\begin{tabular}{|c|c|c|c|c|c|c|c|c|}
\hline
 & & & & & \multicolumn{2}{c|}{ScanNet 10\%} & \multicolumn{2}{c|}{SUN-RGBD 5\%} \\ 
 \cline{6-9} \multirow{-2}{*}{\tabincell{c}{Obj\&Cls\\Filter}} &  
  \multirow{-2}{*}{\tabincell{c}{IoU\\Filter}} &
 \multirow{-2}{*}{\tabincell{c}{Train-time\\Suppression}} & 
 \multirow{-2}{*}{\tabincell{c}{Test-time\\Suppression}} & 
 \multirow{-2}{*}{\tabincell{c}{Test-time\\IoU opt.}} & 
 mAP@0.25 & mAP@0.5 & mAP@0.25  & mAP@0.5 \\ \hline
 & & & Obj-NMS & & 38.4 & 19.8 & 32.9 & 12.5 \\ \hline
\checkmark & & & Obj-NMS & & 44.5 & 24.7 & 36.9 & 17.5 \\ \hline
\checkmark & & Obj-NMS & Obj-NMS & & 44.2 & 25.2 & 37.1 & 17.4 \\ \hline \hline
\checkmark & \checkmark & IoU-NMS & Obj-NMS & & 45.9 & 26.8 & 37.4 & 18.7 \\ \hline
\checkmark & \checkmark & IoU-LHS & Obj-NMS & & 46.5 & 26.9 & 37.9 & 18.5 \\ \hline
\checkmark &  \checkmark & IoU-LHS & IoU-NMS & & 47.0 & 28.2 & 38.8 & 20.8 \\ \hline

\checkmark & \checkmark & IoU-LHS & IoU-NMS & \checkmark & \textbf{47.2} & \textbf{28.3} & \textbf{39.0} & \textbf{21.1} \\ \hline
\end{tabular}
\vspace{1mm}
\caption{Effects of the different components, including train-time filtering and deduplication, and test-time improvements.}
\label{tab:ablation}
\end{table*}

\subsection{Experiments on Indoor Scene Datasets}
For experiments on indoor datasets, \textit{i.e.}, ScanNet and SUNRGB-D, we use IoU-aware VoteNet as our backbone detector.

\subsubsection{Result Comparison}

Table \ref{tab:main} shows the results of our method compared to SESS and VoteNet under different ratios of labeled data on ScanNet and SUN RGB-D, respectively. The results illustrate that, with our effective train-time filtering and test-time improvement leveraging IoU estimation, we are able to significantly outperform current state-of-the-art, SESS, under all labeled ratio settings. With 5\% labeled
data, our method outperforms SESS by \textbf{8.1} and \textbf{8.0} on mAP@0.5 on ScanNet and SUN RGB-D, respectively. Note that our method gains more improvement on mAP@0.5, thanks to the high quality of pseudo labels and the IoU guidance for test-time NMS. 


\subsubsection{Ablation Study}
\paragraph{Filtering and Deduplication Mechanism.}
We study the effect of each component of the filtering and deduplication mechanism. In Table \ref{tab:ablation}, the second row shows the results of naive pseudo labeling, which takes all predictions from the teacher model for supervision. Expectedly the results are not satisfying, only a little higher than VoteNet. Simply applying the dual filtering of classification and objectness confidence gives significant improvement, as the filtering picks out the teacher model proposals that are very likely to be close to true objects and have the correct class. The conventional objectness-based NMS in VoteNet, however, fails to improve further, since the remaining proposals already have high objectness scores and the objectness-based NMS is not capable of picking the ones with higher localization accuracy.

As shown in the fifth and sixth row, after we introduce IoU during train time, IoU filtering and train-time IoU-guided NMS contribute to better performance under both settings. Our proposed IoU-guided LHS improves over IoU guided NMS on mAP@0.25, since LHS finds a better balance between quality and coverage. With better filtering and deduplication leveraging IoU estimation during train time, we gain 2.3 and 1.7 absolute improvement over the without-IoU version on mAP@0.25 and mAP@0.5 respectively on ScanNet 10\%. This verifies that considering localization confidence is important for getting high-quality pseudo labels. With test-time improvements, our method gains in total 3.0 and 3.1 absolute improvement respectively.

\begin{figure}[t]
\begin{center}
   \includegraphics[width=1.0\linewidth]{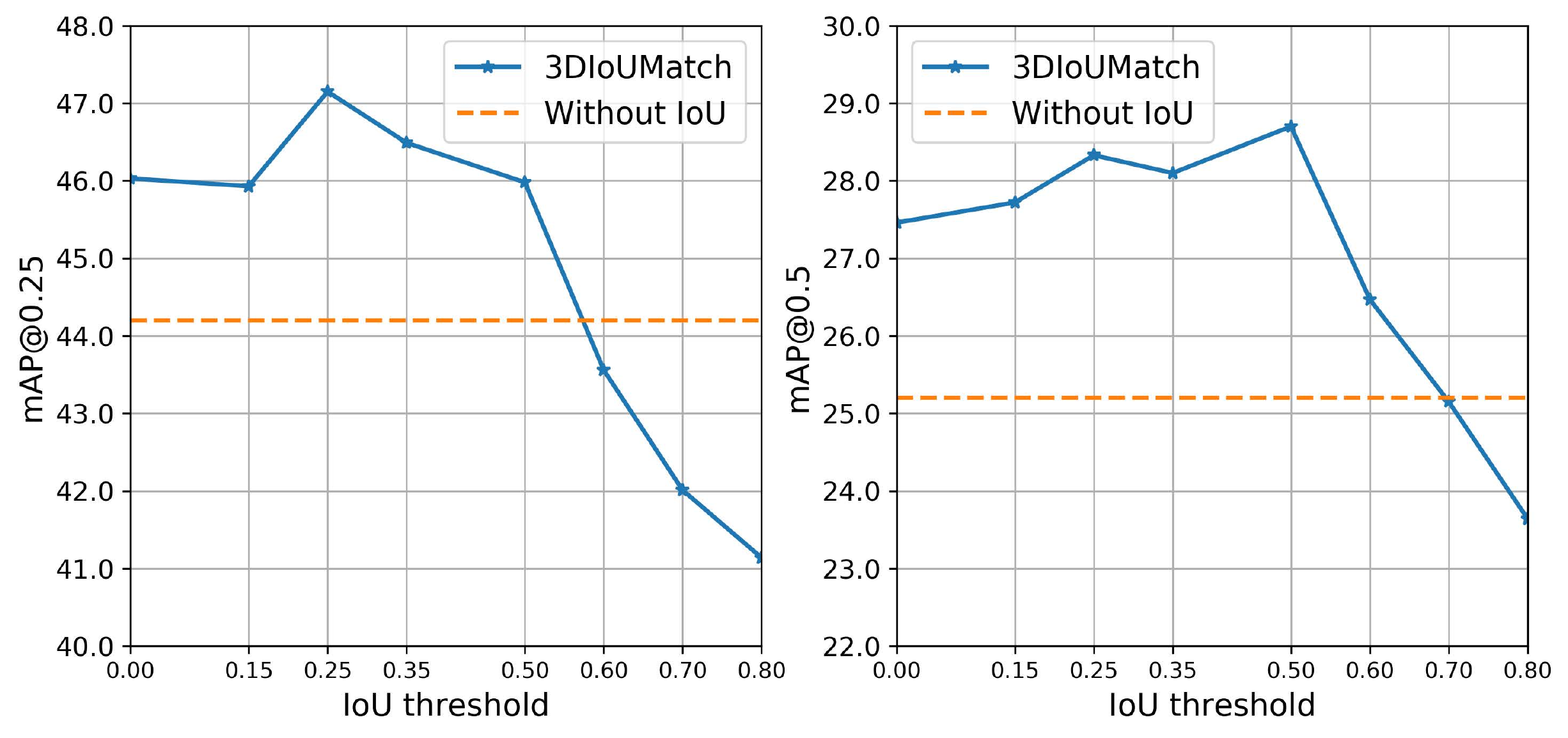}
\end{center}
   \caption{3DIoUMatch results with different IoU thresholds on ScanNet 10\%.}
\label{fig:long}
\label{fig:onecol}
\label{fig:iou}
\end{figure}

We set 0.9 for both classification and objectness confidence threshold following STAC~\cite{sohn2020simple} and investigate the effect of different IoU thresholds on ScanNet 10\%, as shown in Figure \ref{fig:iou}. The performance (with test-time improvements) is higher than the without-IoU baseline by large margins when $\tau_{IoU} \leq 0.5$. Note that the performance peaks at $\tau_{IoU} = 0.25$ for mAP@0.25 while peaking at 0.5 for mAP@0.5, simply because mAP@0.5 prefers a stronger filtering on localization quality. When $\tau_{IoU} > 0.5$ , further increasing  $\tau_{IoU}$ may lead to a drastic drop in pseudo-label coverage and hence is detrimental to the performance.

\paragraph{Test-time IoU-guided NMS and IoU optimization.}
We then evaluate the improvement brought by using IoU estimations at test time. The last two rows in Table \ref{tab:ablation} shows that IoU-guided NMS and IoU optimization improves the performance further.

\subsubsection{Result Analysis}
\begin{figure}[t]
\begin{center}
   \includegraphics[width=1.0\linewidth]{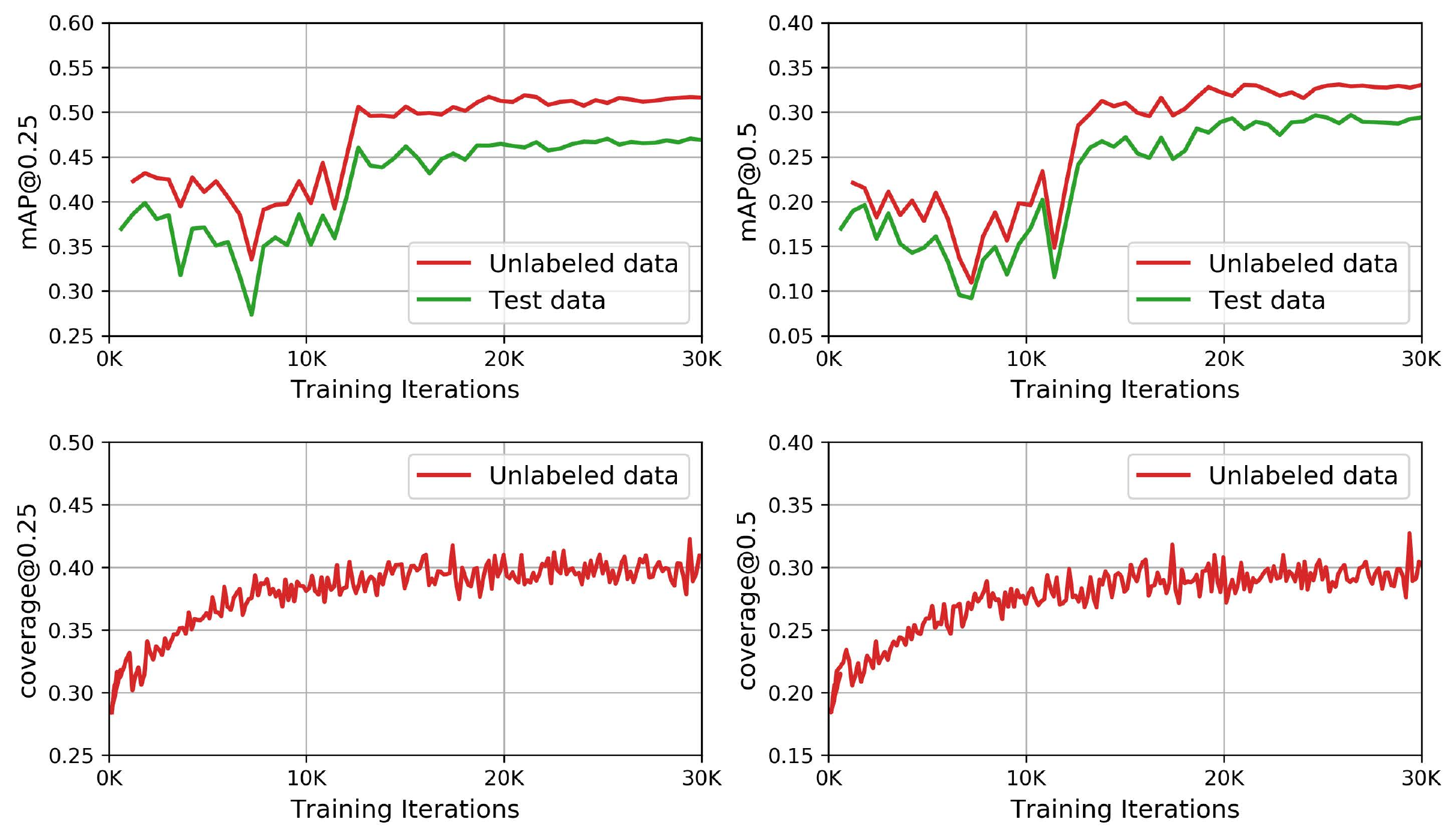}
\end{center}
   \caption{The performance improvements and pseudo-label coverage over the semi-supervised learning stage on ScanNet 10\%.}
\label{fig:long}
\label{fig:onecol}
\label{fig:stats}
\end{figure}

In this section, we examine how our 3DIoUMatch works during training on ScanNet 10\%. The upper two curves in Figure \ref{fig:stats} show that as the training goes, the performance on unlabeled data and test data increases conformably, which indicates the increasing quality of pseudo-labels. We also show how the coverage of the pseudo-labels changes on the unlabeled data over the training. Here coverage at a certain threshold simply means the class-agnostic recall, measuring the percentage of ground truth objects that can find a pseudo-label with an IoU larger than the threshold.
As we can see from the lower two curves in Figure \ref{fig:stats}: at the beginning, the coverage of the pseudo-labels is relatively low due to the strict filtering mechanism; as the semi-supervised learning goes on, the improving detection performance leads to a higher passing rate of the filter and hence a higher coverage of the pseudo-labels, which in return fuels SSL; by the end of training, the coverage at 0.25 and at 0.5 both increase by about 10\%. 

\subsubsection{Implementation Details}
\mypara{Training}
For the pre-training stage, we train with a batch size of 8 and follow the same data augmentation of SESS~\cite{zhao2020sess}.  We then use those pre-trained weights to initialize the student and teacher networks. 
For the SSL stage, we construct each batch by taking 4 labeled samples and 8 unlabeled samples, with the same data augmentation. The weights of different loss terms are the same as VoteNet and we set $\lambda_{u}=2$. The student network is trained for 1000 epochs (the labeled data is traversed in one epoch), optimized by an ADAM optimizer with an initial learning rate of 0.002, and the learning rate is decayed
by 0.3, 0.3, 0.1, 0.1 at the 400\textsuperscript{th}, 600\textsuperscript{th}, 800\textsuperscript{th} and 900\textsuperscript{th} epoch, respectively. The number of generated 3D proposals is 128. We use $k=3, D=4$ for the IoU module. The three thresholds are set to be $\tau_{obj}=0.9, \tau_{cls}=0.9, \tau_{IoU}=0.25$. 
For more details, we refer the readers to the supplementary materials.

\mypara{Inference}
We forward the input to the student network to generate proposals. We first apply IoU optimization to refine box parameters following IoU-Net~\cite{jiang2018acquisition}, followed by an IoU-guided NMS with a 3D IoU threshold of 0.25.


\subsection{Experiments on KITTI}\label{sec:kitti}
For experiments on the KITTI dataset,  we adopt PV-RCNN\cite{shi2020pv} as our backbone\footnote{Update to the CVPR camera-ready version: to avoid information leakage, we no longer apply the ground truth sampling augmentation used by PV-RCNN to unlabeled data in 1\% and 2\% labeled data experiments. We therefore retrain the networks and update the results in Table 3 and 4. See supplementary materials for more details.}. PV-RCNN itself comes with a 3D IoU module, allowing to use it in our semi-supervised learning pipeline without modifying its architecture.
\vspace{-3mm}
\subsubsection{Results}
We evaluate our method on KITTI validation set. Table \ref{tab:kitti_main} demonstrates significant and consistent improvement across all categories with 1\%, 2\%, and 100\% labeled data, compared to supervised training only. Similar to our experiments on indoor scene datasets, here the 100\% labeled data setting means making a copy of the full dataset as unlabeled data and training the network using our devised semi-supervised pipeline. With 2\% labeled data, our method outperforms the labeled-data-only baseline by 7.4 and 10.7 mAP@0.5 on pedestrian and cyclist, respectively, which are very challenging for low data ratios.

\vspace{-3mm}
\subsubsection{Ablation Study}
We conduct ablation studies on KITTI with 1\% labeled data. Table \ref{tab:kitti_ablate} shows our improvements compared with a pseudo-label baseline and filtering based on class confidence only. The results validate the effectiveness of our IoU-based localization confidence filtering.

\vspace{-3mm}

\subsubsection{Implementation Differences with VoteNet}

In KITTI, we only care about three classes, car, pedestrian, and cyclist, which differ a lot in the difficulty to detect. 
Instead of using LHS, we follow PV-RCNN to set different IoU thresholds for each individual class, \textit{i.e.},
$\tau_{car}= 0.5$, $\tau_{ped}=\tau_{cyc} = 0.25$. 
In contrast to VoteNet, PV-RCNN is a two-stage detector containing an RPN. After the RPN stage, only a number of proposals are kept based on classification score, which acts like objectness score in VoteNet. We further filter with the classification score obtained from the RPN stage, with a threshold of $\tau_{cls}=0.4$.

Due to the non-differentiability of the IoU module of PV-RCNN, we do not apply IoU optimization.

We also adopt a two-stage training scheme for our PV-RCNN-based 3DIoUMatch. We use an unlabeled weight $\lambda_{u}=1$ and only supervise anchor classification and bounding box regression in PV-RCNN on unlabeled data. Please refer to the supplementary materials for more details.

\begin{table}[t]
 \resizebox{\columnwidth}{!}{
      \centering
    \begin{tabular}{|c|c|c|c|c|c|c|c|c|c|}
    \hline
    & \multicolumn{3}{c|}{1\%} & \multicolumn{3}{c|}{2\%} & \multicolumn{3}{c|}{100\%}\\ 
      \cline{2-10} \multirow{-2}{*}{} & Car & Ped. & Cyc. & Car & Ped. & Cyc. & Car & Ped. & Cyc.  \\ \hline
    PVR. & 73.5 & 28.7 & 28.4 & 76.6 & 40.8 & 45.5 & 83.0 & 57.9 & 73.1 \\ \cline{1-10} 
    Ours & \textbf{76.0} & \textbf{31.7} & \textbf{36.4} & \textbf{78.7} & \textbf{48.2} & \textbf{56.2} & \textbf{84.8} & \textbf{60.2} & \textbf{74.9} \\ \cline{1-10} 
    Improvements & \textbf{+2.5} & \textbf{+3.0} & \textbf{+8.0} & \textbf{+2.1} & \textbf{+7.4} & \textbf{+10.7} & \textbf{+1.8} & \textbf{+2.3} & \textbf{+1.8} \\ \cline{1-10} 
    \end{tabular}
    }
\vspace{1mm}
    \caption{\textbf{3D detection results on KITTI val set with different labeled ratios.} The results are for moderate difficulty level evaluated by the mAP with 40 recall positions, with a rotated IoU threshold 0.7, 0.5, 0.5 for the three classes, respectively.}
    \label{tab:kitti_main}
    
\end{table}

\begin{table}[t]
 \resizebox{\columnwidth}{!}{
      \centering
\begin{tabular}{|c|c|c|c|c|c|c|c|c|c|}
\hline
& \multicolumn{3}{c|}{Car} & \multicolumn{3}{c|}{Pedestrian} & \multicolumn{3}{c|}{Cyclist}\\ 
  \cline{2-10} \multirow{-2}{*}{} & Easy & Mod. & Hard & Easy & Mod. & Hard & Easy & Mod. & Hard  \\ \hline
PVR. & 87.7 & 73.5 & 67.7 & 32.4 & 28.7 & 26.2 & 48.1 & 28.4 & 27.1 \\ \cline{1-10} 
naive psd.-lb. & 88.4 & 75.2 & 69.5 & 32.7 & 29.2 & 26.7 & 51.4 & 30.7 & 28.7 \\ \cline{1-10}
cls. filt. only & 87.9 & 75.5 & 70.5 & 36.6 & 31.0 & 28.3 & 57.3 & 35.3 & 33.0 \\ \cline{1-10} 
Ours & \textbf{89.0} & \textbf{76.0} & \textbf{70.8} & \textbf{37.0} & \textbf{31.7} & \textbf{29.1} & \textbf{60.4} & \textbf{36.4} & \textbf{34.3} \\ \hline
\end{tabular}
    }
\vspace{1mm}
\caption{\textbf{Ablation study on KITTI 1\% labeled data. }Same evaluation metric as Table 1. 
}
\label{tab:kitti_ablate}
\vspace{-2mm}

\end{table}

\section{Conclusion}
In this paper, we propose 3DIoUMatch, a novel semi-supervised 3D object detection method leveraging IoU estimation. Built upon a teacher-student mutual learning framework, we leverage asymmetric data augmentation and pseudo-label filtering and deduplication to facilitate the student learning from the EMA teacher. With our IoU estimation module, we make filtering and deduplication aware of localization confidence and apply test-time IoU-guided NMS and IoU optimization, leading to further improvement. Experiment results on the ScanNet, SUN-RGBD, and KITTI datasets validate the effectiveness of our method: we achieve significant gain over the previous state-of-the-art methods and baselines under all settings. Our idea of leveraging IoU estimation for semi-supervised learning is generally applicable to different kinds of 3D object detectors and can be extended to 2D detectors as future works.

{
\vspace{2mm}
\noindent \textbf{Acknowledgement:}
This research is supported by a grant from the SAIL-Toyota Center for AI Research, NSF grant CHS-1528025, a Vannevar Bush Faculty fellowship, a TUM/IAS Hans Fischer Senior Fellowship, and gifts from the Adobe, Amazon AWS, and Snap corporations.}

   

\appendix
\section{3D IoU Estimation Module for VoteNet}\label{sec:iou-module}
To facilitate the rejection of poorly localized proposals, as well as guiding deduplication and test-time refinement, we devise a new 3D IoU estimation module differentiable w.r.t bounding box parameters for point-cloud-based IoU-unaware detectors like VoteNet.

In detail, for each predicted bounding box $b^{(k)}$, we wish to estimate its 3D IoU $v^{(k)} \in [0,1]$ with respect to its corresponding ground-truth box $\{ {o^k \in \{o^{(j)}\}} | k = \text{argmax}_j(\text{IoU}(b^{(k)}, o^{(j)}) )\}$. VoteNet does not have intermediate region proposals and only output bounding box parameters at the end stage. Features used for bounding box parameter regression are gathered from vote points in a fixed-radius ball vicinity around each vote cluster, which are unaware of the final bounding box prediction. So, different from implementation in IoU-Net~\cite{jiang2018acquisition} that parallel the bounding refinement and IoU estimation, we need to do it serially by pooling features again specifically for 3D IoU estimation using the final predicted bounding box.

This feature pooling layer takes a bounding box as input and should generate continuous features with respect to the change in bounding box parameters. Existing RoI pooling methods proposed in GSPN~\cite{yi2019gspn} and PointRCNN~\cite{shi2019pointrcnn} and 3D IoU module proposed in STD~\cite{yang2019std} simply set a hard cropping boundary at the bounding box surface, taking the point features inside the proposal and ignoring any points outside. These designs have poor differentiability and cause discontinuities whenever a change in the box parameters modifies the point population inside the box, thus are not suitable for 3D IoU optimization (see Table \ref{tab:iou-effect}).


Here, for the first time, we devise a 3D pooling layer, 3D Grid Pooling, that is differentiable with respect to the change in all bounding box parameters. 
Inspired by RoIAlign~\cite{he2017mask} in 2D object detection, we propose to construct virtual grid points spanning the space of the bounding box and their features are obtained by distance-weighted interpolation from real points not restricted inside the box.


\mypara{Network architecture for IoU-aware VoteNet}
Taking as inputs the seed points $\{z_i\}_{i=1}^{M}$, predicted bounding box $b =\{\mathbf{c}, \mathbf{d}, \theta\}$, and a predicted label $l$, our 3D IoU module estimates the largest 3D IoU between $B$ and all ground truth bounding boxes. Following IoU-Net~\cite{jiang2018acquisition}, the IoU estimation is class-aware.

To build a differentiable 3D IoU module, we first construct a $D\times D \times D$ grid $\{g_m \in \mathbb{R}^{3} ~|~m\in [0, D^3 -1]\}$ that exactly span over the space of $b$ and evenly divide its width, length, and height. For each grid point $g_m$, we find its $k$ nearest neighbours among all seed points and interpolate their features to get $f_m = \frac{\Sigma_{i=1}^{k}w_i f_i}{\Sigma_{i=1}^{k} w_i} $, where $w_i=\frac{1}{d(g_m, g_i)^2}$ and $d$ is the L2 distance. Ideally, if $k$ is equal to the number of all seed points, then the IoU module is continuously differentiable. Due to the computational cost, we empirically find $k=3$ is sufficient for accurate 3D IoU estimation with smooth gradients. We then concatenate $g_m$ and $f_m$ for each grid point and form a grid feature set $\{[g_m;f_m]\}$. The feature set will be pushed towards a PointNet to predict class-aware 3D IoU. A final 3D IoU selection will be performed using the input class label.

Our IoU-aware VoteNet shares the same structure with VoteNet\cite{qi2019deep} except for the IoU estimation module. We provide a more detailed description of the IoU estimation module here. The IoU estimation module is appended after the proposal generation module of VoteNet and takes the bounding box proposals as input. For each bounding box proposal, we create $4\times 4\times 4$ virtual grid points. We obtain the relative coordinates of the grid points by subtracting the coordinates of the bounding box center. For every grid point we find its $k$ nearest neighbours among all seed points and interpolate their features to get $f_m = \frac{\Sigma_{i=1}^{k}w_i f_i}{\Sigma_{i=1}^{k} w_i} $, where $w_i=\frac{1}{d(g_m, g_i)^2}$ and $d$ is the L2 distance. The interpolated features of every grid point is then concatenated with the relative coordinates and forwarded into an MLP with channel dimensions of [256+3, 128, 128, 128] to learn a new feature. Then the features of all grid points go through a 
global max pooling, after which go through another MLP with channels [128, 128, 128, C], where $C$ is the number of classes, to make the IoU prediction class-aware. Finally, we select the per box IoU estimation by using the class label (during training) or class prediction (during inference).

\mypara{Training IoU Estimation Module}
To train the 3D IoU estimation branch in both stages, we generate on-the-fly training data via jittering the bounding box predictions, i.e. adding Gaussian noise to the box center and size. As a way of data augmentation, this jittering is essential for the generalization of IoU estimation to unlabeled data. We use an L1 loss to supervise the IoU estimation module. 

\begin{figure}[t]
\begin{center}
   \includegraphics[width=1\linewidth]{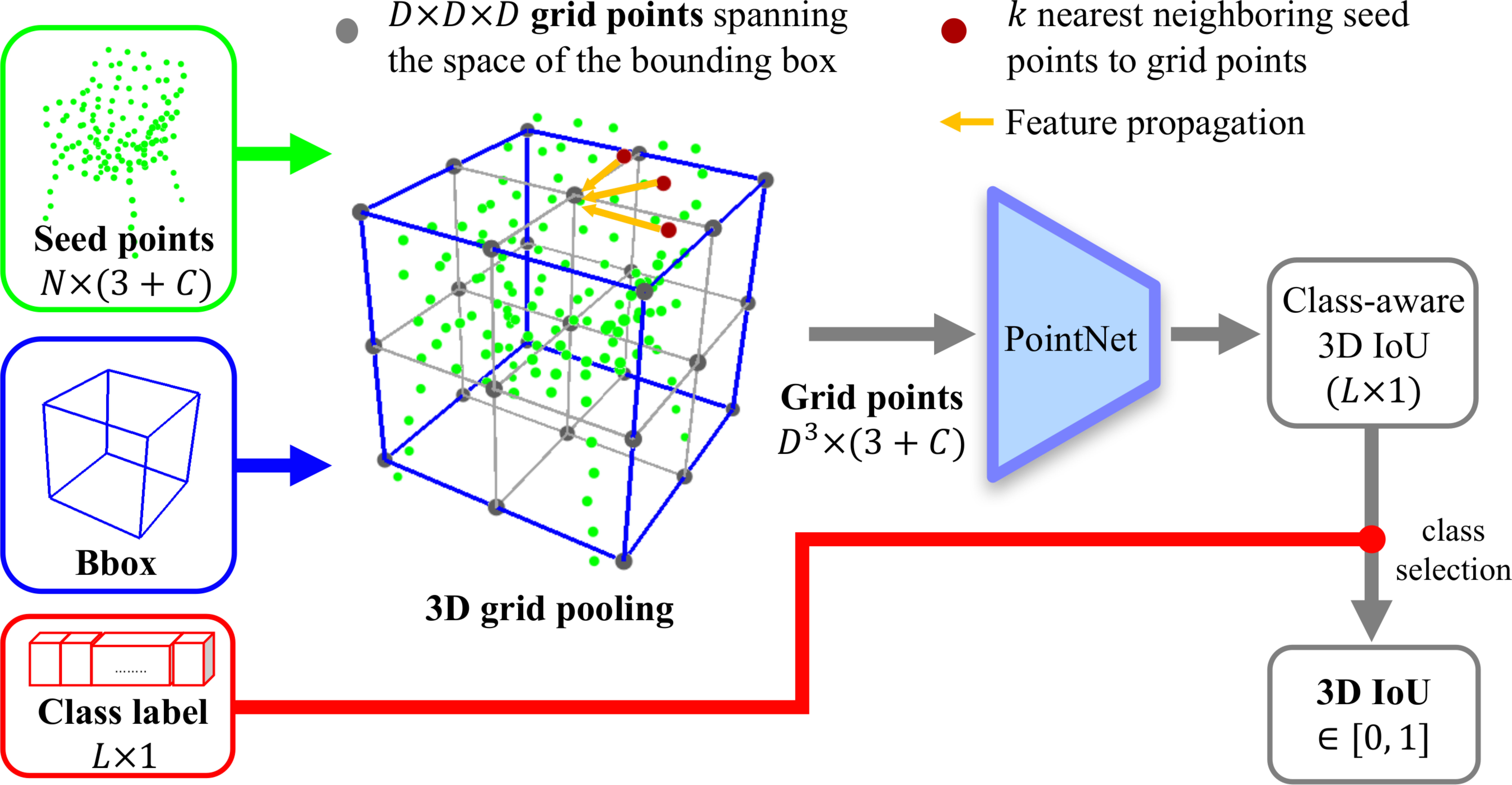}
\end{center}
   \caption{\textbf{3D IoU module} takes inputs seed feature points and a bounding box along with its predicted class label, and estimates the 3D IoU between the box and its maximum overlapping ground truth. 
   The module constructs a 3D regular grid with $D^3$ virtual grid points spanning over the bounding box. We then perform a 3D grid feature pooling that applies a distance-weighted interpolation for feature propagation from the seed points to the grid points. Then the local coordinates of these grid points along with their features are pushed through a PointNet to regress class-aware 3D IoU. Finally, we use the input class label for output selection. }
\label{fig:long}
\label{fig:onecol}
\end{figure}

\section{More Implementation Details for VoteNet-based 3DIoUMatch}

\paragraph{Training}
For the pre-training stage, we find that the network does not converge using the same protocol as fully-supervised VoteNet. We instead use a new protocol, where the network is trained for 900 epochs, optimized by an ADAM optimizer with an initial learning rate of 0.001, and the learning rate is decayed
by 0.1, 0.1, 0.1 at the 400\textsuperscript{th}, 600\textsuperscript{th} and 800\textsuperscript{th}, respectively. We observe convergence using this protocol on all ratios of labeled data. 

Inspired by IoU-Net\cite{jiang2018acquisition}, for both stages, we generate on-the-fly
training data via jittering the bounding box predictions for the IoU estimation module. Specificly, we add $ \epsilon_{size}\sim N(0, (0.3\mathbf{d})^2) $ to each bounding box size prediction $\mathbf{d}$ and add $ \epsilon_{center} \sim N(0, (0.3\mathbf{d})^2) $ to each bounding box center prediction $\mathbf{c}$ to obtain $N_{proposal}$ more training samples. The final IoU estimation loss is the L1 loss averaged over all IoU trainig samples, original predictions or jitters. The IoU estimation loss weight is 1. 

\mypara{Inference}
As IoU-Net\cite{jiang2018acquisition} did not release code, we implemented a simple version of test-time IoU optimization. 
\begin{enumerate}
    \item We obtain the original bounding box proposals.
    \item We calculate the gradients of the IoU estimation w.r.t. to bounding box size and center, $grad_{size}, grad_{center}$, and update the bounding box size and center by adding $grad_{size}\ast \lambda, grad_{center}\ast \lambda$ to the box size and center, respectively, where $\lambda$ is the optimization step size.
    \item We repeat the second step for $T$ times. 
\end{enumerate}

We find setting $T$ to 10 yields noticeable improvement while not slowing inference speed too much. Choosing $\lambda$ from the range of $[1e^{-4}, 5e^{-4}]$ has similar performance.

\section{More Implementation Details for PV-RCNN-based 3DIoUMatch}


We basically follow the training protocols and settings of PV-RCNN~\cite{shi2020pv}. For pre-training on small amounts of labeled data (1\% and 2\%), we lengthen the number of traverses in each epoch to ten times the origin for the model to converge. For SSL training, we set the batch size to 16 (8 labeled + 8 unlabeled, 8 GPUs) and lengthen the number of traverses in each epoch to five times the origin.

PV-RCNN uses a data augmentation method called ground truth sampling. This method first collects all bounding boxes in the whole set of training scenes as well as the point cloud enclosed in the boxes, and then randomly places a sampled set of these boxes and points (called the database) in each scene during training. If a sampled box from the database gets overlapped with an original box in a training scene, this box is discarded. For semi-supervised experiments in which we only use a small fraction of labels, we have to downsize the database to contain boxes from the exact set of used labels only. For pseudo-label generation, we even should not use this augmentation since we have no annotation, thus failing to remove sampled boxes that overlap with ground truth objects.

During training, our pseudo-labels are generated by the teacher model in train mode. However, with limited labels, we should not leverage ground truth when selecting RPN proposals to feed to the second stage. Therefore, we adopt the test-time RPN proposal selection strategy of PV-RCNN when it comes to generating pseudo-labels under a semi-supervised setting.

Nevertheless, for 100\%-label setting, we can leverage all labels, so we are free to use train mode for pseudo-label generation, and we adopt ground truth sampling augmentation for the "unlabeled" data. We also set thresholds to be $\tau_{car}= 0.8, \tau_{ped}=\tau_{cyc} = 0.4, \tau_{cls}=0.2$ for 100\% data, since the pre-trained model of 100\% data should be more accurate in predicting the position of bounding boxes. In the second stage of 100\%-label experiment, we train for 40 epochs, without lengthening each epoch.

\section{Overhead of the IoU module}

Our light-weighted IoU estimation module brings moderate overhead to the network, as shown in Table \ref{tab:train_speed}. The memory reported in the second column refers to the memory consumed by training with batch size 8 on a single GTX 1080Ti GPU. The last two columns mean the time consumed by a full pass (forwarding and backwarding) of a batch of 8 on a single GTX 1080Ti GPU, training ScanNet and SUNRGB-D respectively. Note that regardless of the network design, there is overhead introduced by calculating the ground truth IoU for supervision.

\begin{table}[h]
    \centering
    \begin{tabular}{|c|c|c|c|}
    \hline
        Method & Mem. (GB) & ScanNet (s) & SUNRGB-D (s)  \\ \hline \hline
        VoteNet & 6.56 & 0.282 & 0.316 \\ \hline
        Ours & 6.60 & 0.325 & 0.377 \\ \hline
    \end{tabular}
    \vspace{1mm}
    \caption{Memory and time overhead of the IoU module.}
    \label{tab:train_speed}
\end{table}

\begin{table*}[h]
\begin{adjustbox}{width=\textwidth}
\begin{tabular}{ccccccccccccccccccc}
\hline
& \multicolumn{1}{l}{cab} & \multicolumn{1}{l}{bed} & \multicolumn{1}{l}{chair} & \multicolumn{1}{l}{sofa} & \multicolumn{1}{l}{table} & \multicolumn{1}{l}{door} & \multicolumn{1}{l}{wind} & \multicolumn{1}{l}{bkshf} & \multicolumn{1}{l}{pic} & \multicolumn{1}{l}{cntr} & \multicolumn{1}{l}{desk} & \multicolumn{1}{l}{curt} & \multicolumn{1}{l}{fridg} & \multicolumn{1}{l}{showr} & \multicolumn{1}{l}{toil} & \multicolumn{1}{l}{sink} & \multicolumn{1}{l}{bath} & \multicolumn{1}{l}{ofurn} \\ \hline\hline
VoteNet mAP@0.25     & 17.9                        & 74.7                    & 74.5                      & 75.3                     & 45.6                      & 18.3                     & 11.7                       & 21.7                          & 0.7                         & 28.4                        & 49.4                     & 21.5                        & 23.2                             & 18.5                               & 79.6                       & 25.7                     & 66.3                        & 11.7                           \\ \hline

SESS mAP@0.25        & 20.5                        & 75.1                    & 76.2                      & 76.4                     & 48.1                      & 20.0                     & 14.4                       & 19.4                          & 1.2                         & 30.0                        & 51.8                     & 25.0                        & 30.0                             & 26.4                               & 82.2                       & 29.2                     & 72.3                        & 14.1                           \\ \hline

Without IoU mAP@0.25 & 22.6                        & 79.5                    & 77.8                      & 77.8                     & 49.6                      & 25.4                     & 18.6                       & 27.7                          & 3.3                         & 41.4                        & 56.2                     & 27.4                        & 30.4                             & \textbf{53.6}                      & 81.3                       & 28.5                     & 74.5                        & 18.8                           \\ \hline

3DIoUMatch mAP@0.25  & \textbf{26.6}               & \textbf{82.6}           & \textbf{80.9}             & \textbf{83.3}            & \textbf{52.1}             & \textbf{28.0}            & \textbf{19.9}              & \textbf{29.4}                 & \textbf{3.7}                & \textbf{45.0}               & \textbf{61.9}            & \textbf{29.2}               & \textbf{34.1}                    & 51.2                               & \textbf{85.7}              & \textbf{32.3}            & \textbf{82.8}               & \textbf{21.5}                  \\ \hline\hline
VoteNet mAP@0.5      & 3.2                         & 64.6                    & 43.4                      & 49.3                     & 25.1                      & 2.8                      & 1.1                        & 8.7                           & 0.0                         & 2.4                         & 14.7                     & 3.9                         & 7.6                              & 1.1                                & 46.8                       & 11.9                     & 39.4                        & 1.5                            \\ \hline
SESS mAP@0.5         & 3.7                         & 61.2                    & 48.0                      & 44.8                     & 29.5                      & 3.2                      & 2.8                        & 8.4                           & 0.2                         & 7.5                         & 19.2                     & 5.0                         & 12.2                             & 1.8                                & 48.7                       & \textbf{15.3}            & 40.8                        & 2.2                            \\ \hline
Without IoU mAP@0.5  & 3.9                         & 66.1                    & 52.7                      & 50.7                     & 35.1                      & 7.9                      & 5.0                        & 13.1                          & \textbf{0.9}                & 14.5                        & 26.1                     & \textbf{10.3}               & 17.5                             & \textbf{7.0}                       & 63.9                       & 11.7                     & 62.1                        & 4.9                            \\ \hline
3DIoUMatch mAP@0.5   & \textbf{5.9}                & \textbf{72.0}           & \textbf{60.5}             & \textbf{56.6}            & \textbf{39.7}             & \textbf{10.3}            & \textbf{5.2}               & \textbf{18.1}                 & 0.7                         & \textbf{16.0}               & \textbf{35.3}            & 8.3                         & \textbf{21.4}                    & 6.2                                & \textbf{67.5}              & 13.2                     & \textbf{67.6}               & \textbf{5.2}                   \\ \hline
\end{tabular}
\end{adjustbox}
\caption{Per class mAP@0.25 and mAP@0.5 on ScanNet val set, with 10\% labeled data.}
\label{tab:scanperclass}
\end{table*}

\begin{table*}[h]
\begin{adjustbox}{width=\textwidth}
\begin{tabular}{ccccccccccc}
\hline
\multicolumn{1}{c}{} & \multicolumn{1}{c}{bathtub} & \multicolumn{1}{c}{bed} & \multicolumn{1}{c}{bookshelf} & \multicolumn{1}{c}{chair} & \multicolumn{1}{c}{desk} & \multicolumn{1}{c}{dresser} & \multicolumn{1}{c}{nightstand} & \multicolumn{1}{c}{sofa} & \multicolumn{1}{c}{table} & \multicolumn{1}{c}{toilet} \\ \hline\hline
VoteNet mAP@0.25     & 67.8                        & 32.2                    & 39.4                          & 58.5                      & 53.5                     & 8.0                         & 1.9                            & 14.7                     & 3.2                       & 20.3                       \\ \hline

SESS mAP@0.25        & 70.8                        & 34.7                    & 41.9                          & 60.4                      & 63.0                     & 9.8                         & 3.7                            & 25.2                     & 4.0                       & 28.0                       \\ \hline

Without IoU mAP@0.25 & 75.1                        & 33.5                    & 43.0                          & 59.5                      & 76.9                     & \textbf{6.8}                & 5.1                            & 33.0                     & 3.5                       & 34.8                       \\ \hline

3DIoUMatch mAP@0.25  & \textbf{75.4}               & \textbf{37.7}           & \textbf{45.2}                 & \textbf{64.2}             & \textbf{77.0}            & 6.0                         & \textbf{5.7}                   & \textbf{34.6}            & \textbf{4.5}              & \textbf{39.4}              \\ \hline\hline

VoteNet mAP@0.5      & 31.2                        & 6.2                     & 15.5                          & 29.6                      & 14.6                     & 0.5                         & 0.2                            & 2.0                      & 0.3                       & 5.2                        \\ \hline
SESS mAP@0.5         & 36.7                        & 7.2                     & 19.2                          & 31.8                      & 20.4                     & 0.7                         & 0.5                            & 7.0                      & 0.4                       & 7.1                        \\ \hline
Without IoU mAP@0.5  & 41.5                        & 9.7                     & 25.7                          & 34.5                      & 40.8                     & \textbf{0.8}                & 0.8                            & 8.3                      & \textbf{0.8}              & 11.4                       \\ \hline
3DIoUMatch mAP@0.5   & \textbf{45.2}               & \textbf{14.4}           & \textbf{27.8}                 & \textbf{43.6}             & \textbf{47.2}            & \textbf{0.8}                & \textbf{1.9}                   & \textbf{15.7}            & 0.6                       & \textbf{13.4}              \\ \hline
\end{tabular}
\end{adjustbox}
\vspace{1mm}
\caption{Per class mAP@0.25 and mAP@0.5 on SUNRGB-D val set, with 5\% labeled data.}
\label{tab:sunperclass}
\end{table*}

\section{IoU Module Comparison}

As mentioned in \ref{sec:iou-module}, some region-based 3D detectors, e.g. STD~\cite{yang2019std}, crop the features inside a predicted bounding box and regress the offset. To capture their core characteristics under IoU optimization, we build a simple IoU estimation module which only queries points inside the predicted box and passes the queried feature points through a PointNet to predict the 3D IoU, namely box query. In principle, the differentiability of this module is the same as that in STD, which doesn't release their code and misses the IoU optimization step in their paper.
For a fair comparison, we train another IoU-aware VoteNet with box query as the IoU estimation module and show the comparison between it and our proposed method on the full set of ScanNet and SUN RGB-D. From the results in Table \ref{tab:iou-effect}, we prove that our method is more effective on both IoU-guided NMS and IoU optimization than box query. 

We provide more explanation on why an IoU estimation network design like that in STD\cite{yang2019std} is less effective in IoU estimation and is not differentiable. Given a bounding box proposal, STD concatenates the canonized coordinates and features of the points inside the bounding box to form new features of the points. Therefore, the new feature of a point $f'$ can be denoted as a function of the point coordinates $\mathbf{p}$, the original point feature $f$, the bounding box center $\mathbf{c}$ and the bounding box heading angle $\theta$. Then STD voxelizes the bounding box and sample points in each voxel to produce the voxel feature $f_v$. The process of producing the voxel feature from points in voxels consist of no other parameters except from the point features $\{f'_i\}$ and point coordinates $\mathbf{p}_i$, so $f_v$ is still a function of $\{f_i\}, \{\mathbf{p}_i\},\mathbf{c},\theta$. As all voxel features are flattened and fed to an MLP, which outputs the final IoU, we can conclude that the IoU estimation is not differentiable w.r.t. bounding box size.

We also argue that for VoteNet, since the number of seed points with features are small (1024), box query methods may have difficulty querying points inside a bounding box, especially if a bounding box is too small. Our method, instead won't suffer from this as we are not confined to points inside the bounding box.

Although STD didn't release code, we still implemented an IoU estimation module according to the paper for better comparison. However, some issues need to be stated. First, since the backbone of STD is very different from VoteNet, the comparison between IoU estimation module alone is inherently problematic. Second, STD aims at outdoor object detection, where the task is slightly different. Third, we adopted most of the parameters of STD in the paper, but changed number of voxels (to 27) and number of points sampled per voxel (to 6) due to memory concerns and the small number of seed points in VoteNet. The results in Table \ref{tab:iou-effect} show the  better performance of our IoU module. We also observe serious overfitting using the STD IoU module, suggesting that it may not be suitable for our problem.

\begin{table}[t]
\begin{tabular}{|c|c|c|c|c|}
\hline
& \multicolumn{2}{c|}{ScanNet} & \multicolumn{2}{c|}{SUN RGB-D}\\ 
  \cline{2-5} \multirow{-2}{*}{Method} & \tabincell{c}{mAP\\@0.25} & \tabincell{c}{mAP\\@0.5} & \tabincell{c}{mAP\\@0.25} & \tabincell{c}{mAP\\@0.5}  \\ \hline
Obj-NMS~\cite{qi2019deep} & 57.84 & 35.99 & 58.01 & 33.44 \\ \cline{1-5} 
\tabincell{c}{Box query\\IoU-NMS only}  & 57.56 & 37.07 & 58.16 & 34.81 \\ \cline{1-5} 
\tabincell{c}{Box query\\IoU-NMS + Optim.} & 57.62 & 37.17 & 58.19 & 34.90 \\ \hline
\tabincell{c}{STD\\IoU-NMS only}  & 57.81 & 36.21 & 58.21 & 34.76 \\ \cline{1-5} 
\tabincell{c}{STD\\IoU-NMS + Optim.} & 57.85 & 36.21 & 58.21 & 34.75 \\ \hline
\tabincell{c}{Ours\\IoU-NMS only} & 57.92 & 37.01 & 58.82 & 36.22 \\ \cline{1-5} 
\tabincell{c}{Ours\\IoU-NMS + Optim.} & \textbf{58.46} & \textbf{37.43} & \textbf{59.11} & \textbf{36.71}  \\ \hline
\end{tabular}
\vspace{1mm}
\caption{Comparison of our IoU module with box-query on ScanNet 100\% and SUN RGB-D 100\% .}
\label{tab:iou-effect}
\end{table}


\begin{figure*}[t]
\begin{center}
   \includegraphics[width=1.0\linewidth]{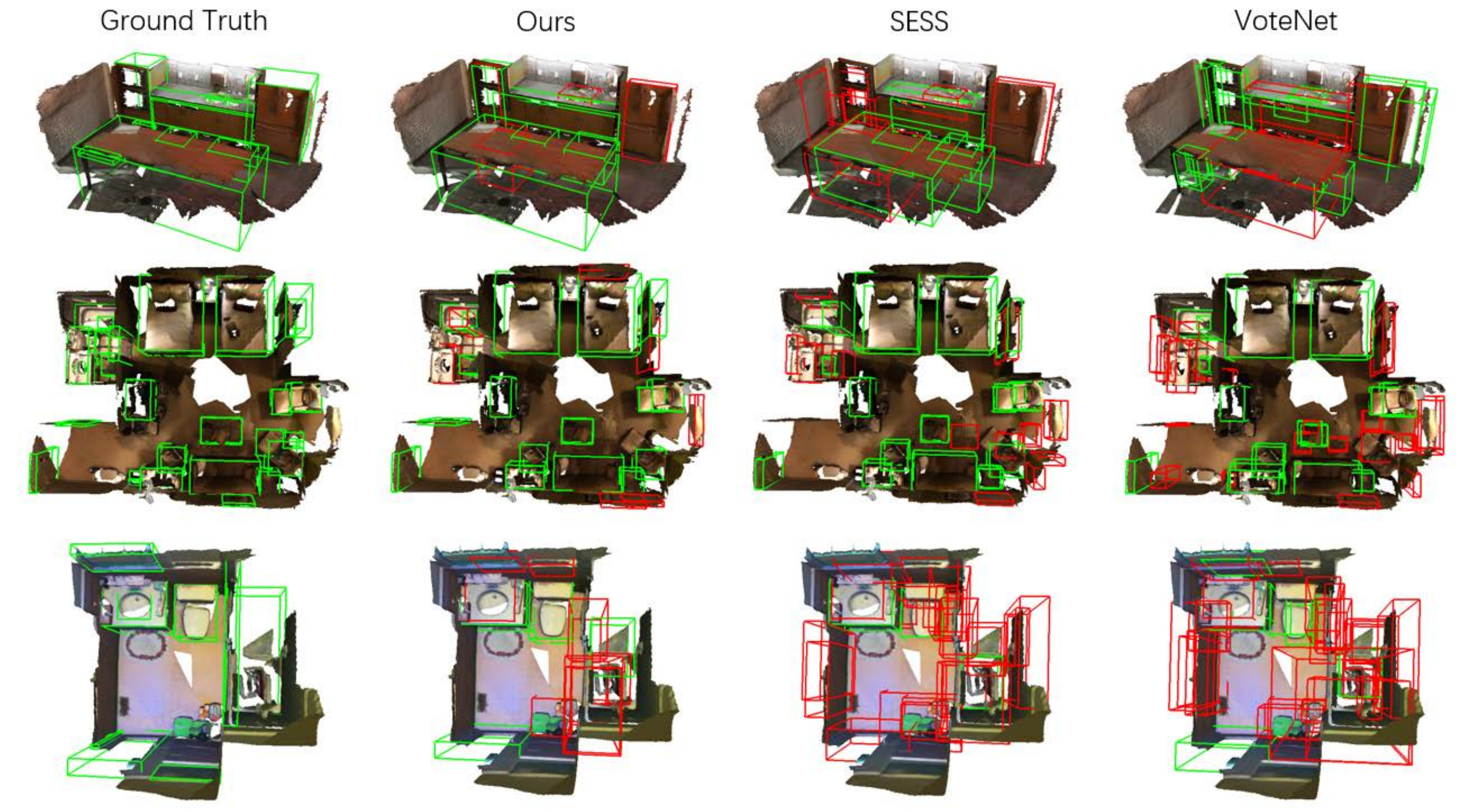}
\end{center}
   \caption{\textbf{Qualitative results on ScanNet, with 10\% labeled data.} Here green bounding boxes have an IoU $\ge 0.25$ while red bounding boxes are with an IoU $<0.25$.}
\label{fig:scan_viz}
\end{figure*}
\begin{figure*}[t]
\begin{center}
   \includegraphics[width=1.0\linewidth]{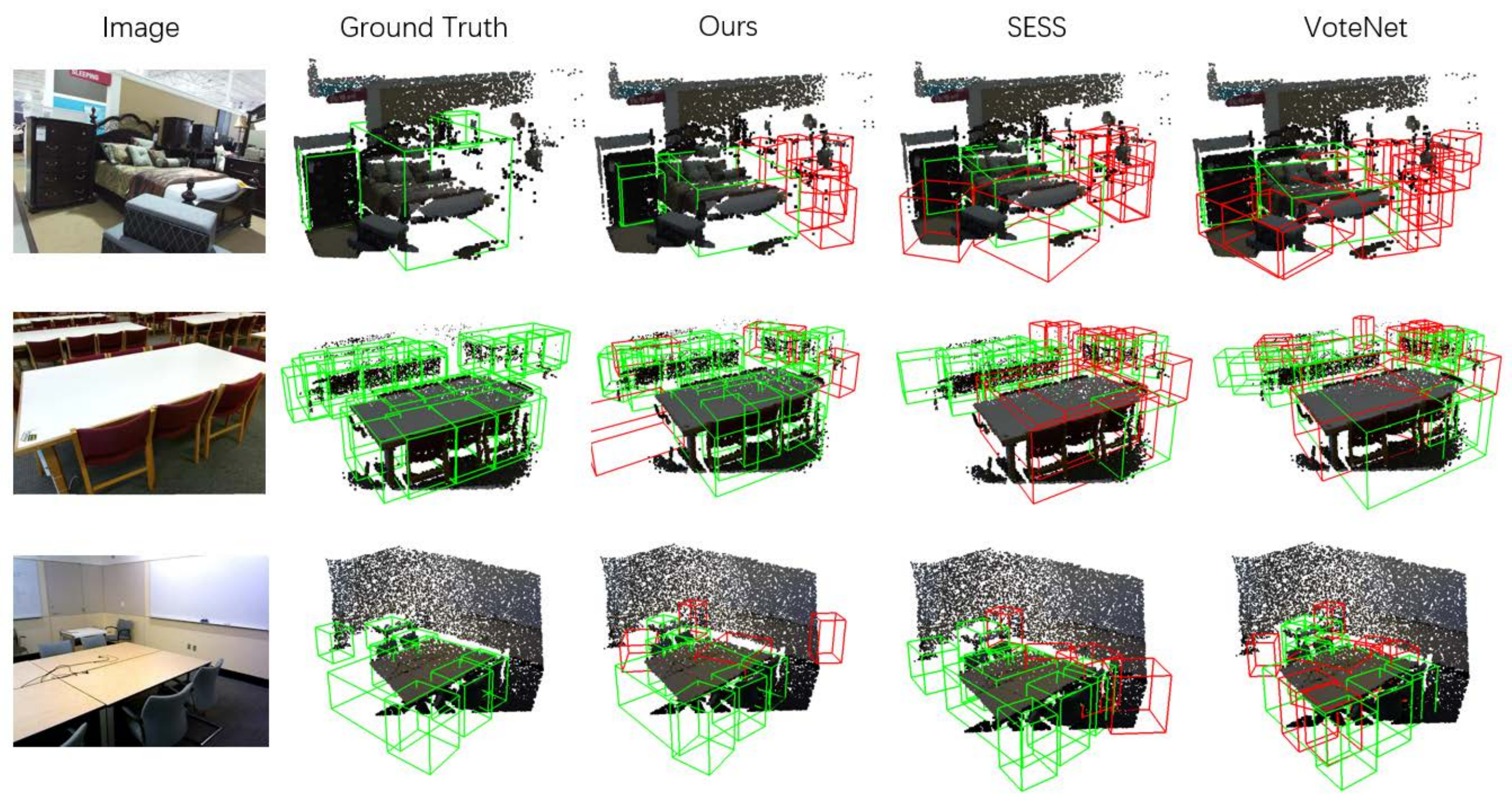}
\end{center}
   \caption{Qualitative results on SUNRGB-D, with 5\% labeled data. }
\label{fig:sun_viz}
\end{figure*}

\section{Why not Supervise Votes and Objectness in VoteNet?}
As we mentioned, we supervise all VoteNet loss terms on unlabeled data except for vote regression loss and objectness binary classification loss.
As we observe, supervising votes or objectness with pseudo labels leads to degrading performance. The main reason is that by rigorous filtering and deduplication we can only be highly confident of a true object being close to a pseudo bounding box, but we are not sure whether or not there is a true object where there are no pseudo bounding boxes nearby. If we supervise objectness on unlabeled data with the pseudo labels the same way as VoteNet, it's not difficult to imagine the network would be more and more biased on detecting objects. In Table \ref{tab:voteobj}, our experiments on ScanNet 10\% and SUNRGB-D 5\% show that the performance suffers a drop after supervising objectness on unlabeled data. 

Vote prediction is an unique component of VoteNet. For a point, the label for its vote is the center of the object it belongs to. To generate pseudo vote labels, the straightforward way is to count every point inside a pseudo bounding box as a vote. However, since this pseudo vote label set is also far from complete, we face a similar problem supervising with it. In Table \ref{tab:voteobj}, our experiments on ScanNet 10\% and SUNRGB-D 5\% also show that the performance drops after supervising vote prediction on unlabeled data.

\begin{table}[h]
    \centering
    \begin{tabular}{|c|c|c|c|c|}
    \hline
          & \multicolumn{2}{c|}{ScanNet 10\%} & \multicolumn{2}{c|}{SUNRGB-D 5\%} \\ \cline{2-5}
        \multirow{-2}{*}{Method} & \tabincell{c}{mAP\\@0.25} &  \tabincell{c}{mAP\\@0.5} &  \tabincell{c}{mAP\\@0.25}& \tabincell{c}{mAP\\@0.5}
        \\ \hline \hline
        3DIoUMatch & 47.2 & 28.3 & 39.0 & 21.1 \\ \hline
        \tabincell{c}{+vote sup.\\ on unlabeled} & 45.4 & 28.3 & 37.9 & 20.9 \\ \hline
        \tabincell{c}{+obj. sup.\\ on unlabeled} & 40.1 & 26.0 & 38.2 & 20.4 \\ \hline
    \end{tabular}
    \vspace{1mm}
    \caption{Objectness \& vote supervision on unlabeled data using pseudo-labels.}
    \label{tab:voteobj}
\end{table}

\section{Per-class Evaluation}


We report per-class average precision on ScanNet with 10\% labeled data and SUNRGB-D with 5\% labeled data, respectively. The bold numbers are the highest per class. The results in Table \ref{tab:scanperclass}, \ref{tab:sunperclass} show that our method improves the average precision on nearly all classes over SESS. Our 3DIoUMatch also has better performance on most classes than the without-IoU version.

\section{Qualitative Results}

We show the qualitative results on ScanNet val set with 10\% labeled training data, Figure \ref{fig:scan_viz} and on SUNRGB-D val set with 5\% labeled training data, Figure \ref{fig:sun_viz}. For the results of our method, SESS and VoteNet, green bounding boxes are the predicted bounding boxes whose IoU $\ge 0.25$, and the red bounding boxes are those with an IoU $<0.25$. As can be seen in both figures, our method give more accurate predictions and significantly reduces the number of false positives.

{\small
\bibliographystyle{ieee_fullname}
\bibliography{egbib}
}

\end{document}